\documentclass[10pt,twocolumn,letterpaper]{article}

\usepackage{cvpr}
\usepackage{times}
\usepackage{epsfig}
\usepackage{graphicx}
\usepackage{booktabs}
\usepackage{amsmath}
\usepackage{amssymb}
\usepackage[linesnumbered,ruled]{algorithm2e}
\usepackage{caption}
\usepackage{subcaption}
\usepackage{multirow}

\graphicspath{ {./results/} }

\usepackage[pagebackref=true,breaklinks=true,letterpaper=true,colorlinks,bookmarks=false]{hyperref}

\cvprfinalcopy 

\ifcvprfinal\fi

\begin{document}
	
	\title{Training deep learning based image denoisers from undersampled measurements
		without ground truth and without image prior}
	
	\author{Magauiya Zhussip, ~~~Shakarim Soltanayev, ~~~Se Young Chun\\
		Ulsan National Institute of Science and Technology (UNIST),
		Republic of Korea \\
		{\tt\small \normalsize \{mzhussip, shakarim, sychun\}@unist.ac.kr}
	}
	
	\maketitle
	\thispagestyle{empty}
	
	\begin{abstract}
	Compressive sensing is a method to recover the original image from undersampled measurements. In order to overcome the ill-posedness of this inverse problem, image priors are used such as sparsity in the wavelet domain, minimum total-variation, or self-similarity. Recently, deep learning based compressive image recovery methods have been proposed and have yielded state-of-the-art performances. They used deep learning based data-driven approaches instead of hand-crafted image priors to solve the ill-posed inverse problem with undersampled data. Ironically, training deep neural networks for them requires ``clean'' ground truth images, but obtaining the best quality images from undersampled data requires well-trained deep neural networks. To resolve this dilemma, we propose novel methods based on two well-grounded theories: denoiser-approximate message passing and Stein's unbiased risk estimator. Our proposed methods were able to train deep learning based image denoisers from undersampled measurements without ground truth images and without image priors, and to recover images with state-of-the-art qualities from undersampled data. We evaluated our methods for various compressive sensing recovery problems with Gaussian random, coded diffraction pattern, and compressive sensing MRI measurement matrices. Our methods yielded state-of-the-art performances for all cases without ground truth images and without image priors. They also yielded comparable performances to the methods with ground truth data.
	\end{abstract}
	
	\begin{figure}[t]
		\begin{center}
			\begin{subfigure}[t]{0.32\linewidth}
				\centering
				\scriptsize Ground Truth
				\includegraphics[width=1\textwidth]{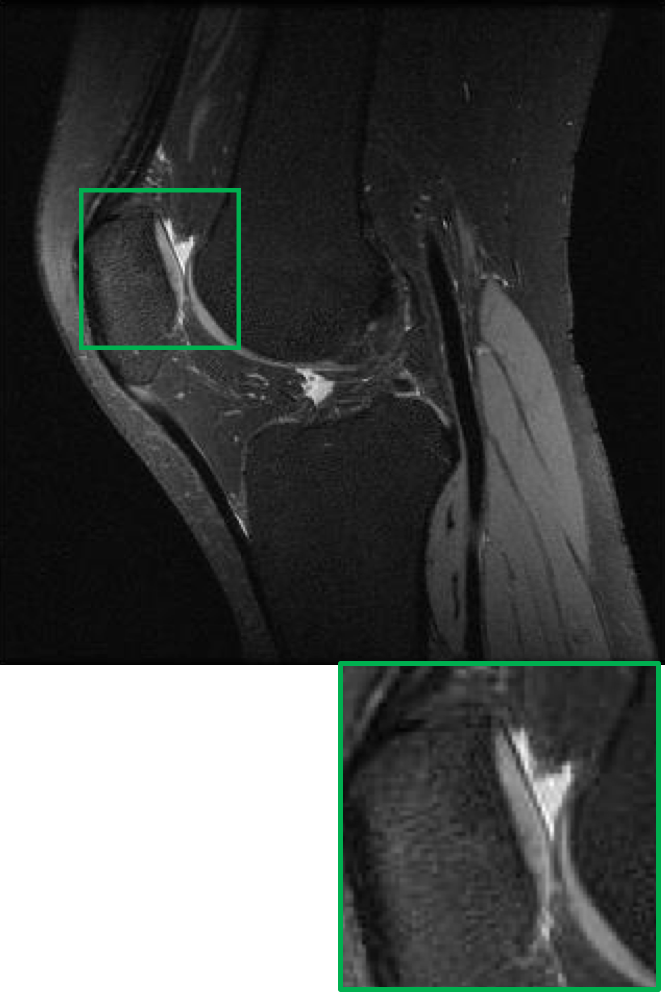}
			\end{subfigure}
			\begin{subfigure}[t]{0.32\linewidth}
				\centering
				\scriptsize TVAL3 (35.74 dB)
				\includegraphics[width=1\textwidth]{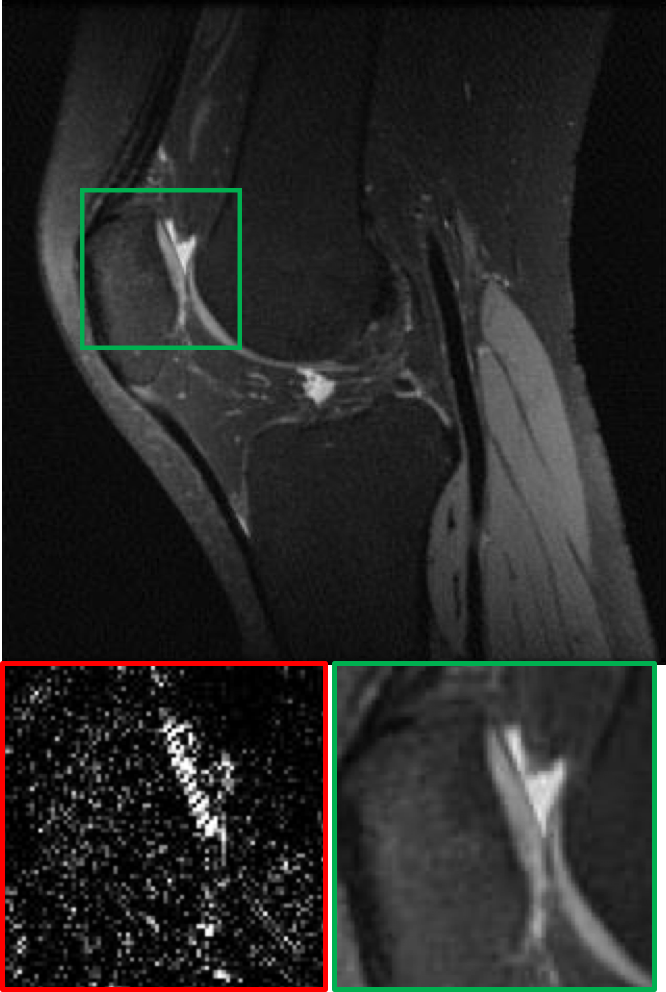}
			\end{subfigure}
			\begin{subfigure}[t]{0.32\linewidth}
				\centering
				\scriptsize Proposed (\textbf{38.67 dB})
				\includegraphics[width=1\textwidth]{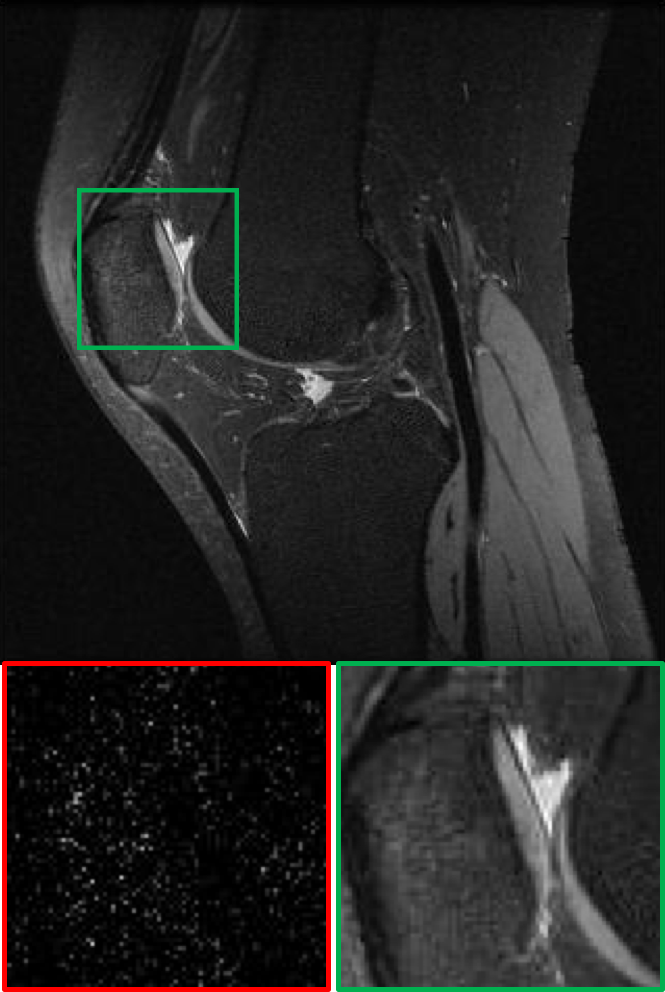}
			\end{subfigure}
		\caption{Ground truth MR image from fully-sampled data (left), reconstructed MR images from 50\%-sampled data using conventional TV image prior (middle,~\cite{Lustig:2007cu}) and our proposed deep learning based method without ground truth (right). Our proposed method yielded significantly better result than conventional method and result comparable to the ground truth with very small residual error (red box).}
		\label{fig-CSMRI_recon50}
		\end{center}
	\end{figure}

	\section{Introduction}
	
    Compressive sensing (CS) has provided ways to sample and to compress signals at the same time with relatively long signal reconstruction time~\cite{Candes:2006eq, Donoho:2006ci}. The idea of combining image acquisition and compression immediately drew great attention in the application areas such as MRI~\cite{Lustig:2007cu,ravishankar2011mr}, CT~\cite{Choi:2010gd}, hyperspectral imaging~\cite{Zhang:2015jn, Zhang:2015ea}, coded aperture imaging~\cite{Arce:2013jm}, radar imaging~\cite{Potter:2010gr} 
	and radio astronomy ~\cite{Wiaux:2009ev}. CS applications have been investigated intensively for the last decade and now some systems are commercialized for practical uses such as low-dose CT and accelerated MR.

	\subsection{Conventional compressive image recovery}
	
	CS is usually modeled as a linear equation:
	\begin{equation}
	{\boldsymbol y} = {\boldsymbol A} {\boldsymbol x} + {\boldsymbol \epsilon}
	\label{eq:lin}
	\end{equation}
	where ${\boldsymbol y} \in \mathbb{R}^M$ is a measurement vector, ${\boldsymbol A} \in \mathbb{R}^{M\times N}$ is a sensing matrix
	with $M \ll N$,
	${\boldsymbol x} \in \mathbb{R}^N$ is an unknown signal to reconstruct, and 
	${\boldsymbol \epsilon} \in \mathbb{R}^M$ is a noise vector. It is a challenging ill-posed inverse problem to estimate $\boldsymbol x$ from the undersampled measurements ${\boldsymbol y}$ with $M \ll N$.
		
	Sparsity has been investigated as  a prior to regularize the ill-posed problem of CS recovery. 
 	CS theories allow to use $l_1$ norm for good signal recovery instead of $l_0$ norm
	~\cite{Candes:2006eq,Donoho:2006ci}, but it requires to use more samples (still significantly lower than $N$). 
	Minimizing $l_1$ norm is advantageous for large-scale inverse problems since $l_1$ norm is convex so that 
	conventional convex optimization algorithms can be used for signal recovery. There have been many convex
	optimization algorithms in solving CS image recovery problem with non-differentiable $l_1$ norm 
	such as iterative shrinkage-thresholding algorithm (ISTA), fast iterative shrinkage-thresholding algorithm 
	(FISTA)~\cite{Beck:2009gh}, alternating direction minimization (ADM)~\cite{Yang:2011dm}, and approximate 
	message passing (AMP)~\cite{Donoho:2009bs}, to name a few. Even though these advanced algorithms 
	achieved remarkable speedup compared to conventional convex optimization algorithms, 
	CS image recovery is still slow in some application areas.

    The image ${\boldsymbol x}$ itself is not usually sparse. However, a transformed image is often sparse. For example, images are sparse in the wavelet domain and/or discrete cosine transform (DCT) domain. In high-resolution imaging, images have sparse edges that are often promoted by minimum total variation (TV)~\cite{li2009user}. Sparse MR image recovery used both wavelet and TV priors~\cite{Lustig:2007cu} or dictionary learning prior from highly undersampled measurements~\cite{ravishankar2011mr}.  Similarly, CS color video and depth recovery used both wavelet and DCT~\cite{Yuan:2014ij}. Hyperspectral imaging utilized manifold-structured sparsity prior~\cite{Zhang:2015jn} or reweighted Laplace prior~\cite{Zhang:2015ea}. Self-similarity in images is also used as a prior for CS image recovery such as NLR-CS~\cite{dong2014compressive} and denoiser based AMP (D-AMP)~\cite{metzler2016denoising}. D-AMP utilized powerful modern denoisers such as BM3D~\cite{dabov2007image} and has recently been extended to sparse MRI~\cite{eksioglu2018denoising}.
	
	\subsection{Deep learning in compressive image recovery}
	
	Deep learning with massive amount of training data has revolutionized many computer vision tasks~\cite{LeCun:2015dt}. It has also influenced many low level computer vision tasks such as image denoising~\cite{burger2012image,jain2009natural,Vincent:2010vu,xie2012image,zhang2017beyond} and CS recovery~\cite{Gupta:2018ia,Hammernik:2017ku, Jin:2017iz,Kulkarni:2016je,metzler2017learned,Xu:2018cd,Zhang:2018wz}.
    There are largely two different approaches using deep neural networks for CS recovery. One is to use a deep network to directly map from an initially recovered low-quality image from CS samples to a high-quality ground truth image~\cite{Jin:2017iz,Kulkarni:2016je}. The other approach for deep learning based CS image recovery is to use deep neural network structures that unfold optimization algorithms and learned image priors, inspired by learned ISTA (LISTA)~\cite{gregor2010learning}.
	In sparse MRI, ADMM-Net~\cite{Yang:2016vc} and variational network~\cite{Hammernik:2017ku} were proposed with excellent performances. Both methods learned parametrized shrinkage function as well as transformation operator for sparse representation from given training data. Recently, 
	instead of using explicit parametrization in shrinkage operator, deep neural networks were used to unfold optimization algorithms for CS image recovery such as 
	learned D-AMP (LDAMP)~\cite{metzler2017learned},
	ISTA-Net~\cite{Zhang:2018wz},
	CNN-projected gradient descent for sparse CT~\cite{Gupta:2018ia}, 
	and Laplacian pyramid reconstructive adversarial network~\cite{Xu:2018cd}.
	Utilizing generative adversarial network (GAN) for CS was also investigated~\cite{Bora:2017tz}.
	All these methods have one important requirement: ``clean" ground truth images must be available for training.

	\subsection{Deep learning without ground truth}
	
All deep learning based approaches for CS image recovery solve the ill-posed inverse problem with undersampled data by using deep neural networks. Ironically, training deep neural networks requires ``clean'' ground truth images, but obtaining the best quality images from undersampled data requires well-trained deep neural networks. It is often expensive or infeasible to acquire clean data, for example, in high-resolution medical imaging (long acquisition time for MR, high radiation dose for CT) or high-resolution hyperspectral imaging. In this paper, we address this dilemma.

	
    Recently, there have been a few attempts to train deep neural networks for low-level computer vision tasks in unsupervised ways. Lehtinen \textit{et al.} proposed noise2noise to train deep neural networks for image denoising, inpainting, and MR reconstruction~\cite{Lehtinen:2018un}. This work implemented MR reconstruction using a direct mapping instead of unfold optimization scheme. However, this was not evaluated with various CS applications and it requires two contaminated data for each image, which may not be available in some cases. Bora \textit{et al.} proposed AmbientGAN, a training method for GAN with contaminated images and applied it to CS image recovery~\cite{Bora:2017tz,Bora:2018tl}. However, AmbientGAN was trained with artificially contaminated images, rather than with CS measurements. Moreover, the method of~\cite{Bora:2017tz} is limited to $i.i.d.$ Gaussian measurement matrix theoretically and was evaluated with relatively low-resolution images. Soltanayev \textit{et al.} proposed a Stein's unbiased risk estimator (SURE) based training method for deep learning based denoisers~\cite{soltanayev2018}. This method requires only one realization, but it is limited to  $i.i.d.$ Gaussian noise. Moreover, it is non-trivial to extend this to CS image recovery.
	
In this paper, 
	we propose unsupervised training methods for deep learning based CS recovery using two theories: D-AMP and SURE. Our proposed methods can train deep denoisers from undersampled measurements without ground truth and without image priors, and to recover images with state-of-the-art qualities. The contributions of our work are:

1) Proposing a method to train deep learning based denoisers from undersampled measurements without ground truth and without image priors. Only one realization for each image was required. An accurate noise estimation method was also developed for training deep denoisers.

2) Proposing a CS image recovery method by modifying LDAMP to have as low as 1 denoiser instead of 9 denoisers with comparable performance to reduce training time.

3) Extensive evaluations of the proposed method using high-resolution natural images and MR images for 3 CS recovery problems with Gaussian random, coded diffraction pattern, and realistic CS-MRI measurement matrices.

	\section{Background}
	
	\subsection{Denoiser-based AMP (D-AMP)}
	
	D-AMP is an algorithm designed to solve CS problems, where one needs 
	to recover image vector ${\boldsymbol x} \in\mathbb{R}^N$ from the measurements ${\boldsymbol y} \in\mathbb{R}^M$ 
	using prior information about ${\boldsymbol x}$. Based on the model (\ref{eq:lin}), the problem can be formulated as:
	\begin{equation} 
	\min_{\boldsymbol x} \|{\boldsymbol y}-{\boldsymbol A}{\boldsymbol x}\|^2_2 \quad \textrm{subject to} \quad {\boldsymbol x} \in C \label{eq1}
	\end{equation}
	where $C$ is a set of natural images. 
	D-AMP solves (\ref{eq1}) relying on AMP theory. 
	It employs appropriate Onsager correction term ${\boldsymbol b_t}$ at each iteration, 
	so that ${\boldsymbol x}_t + {\boldsymbol A}^H {\boldsymbol z}_t$ in Algorithm~\ref{algorithmDAMP} becomes close to 
	the ground truth image plus $i.i.d.$ Gaussian noise. 
	D-AMP can utilize 
	any state-of-the-art denoiser as a mapping operator ${\boldsymbol D}_{{\boldsymbol w}(\hat{\sigma_t})}( \cdot )$
	in CS image recovery (Algorithm~\ref{algorithmDAMP})
	for reducing $i.i.d.$ Gaussian noise as far as the divergence of denoiser can be obtained.
	
	\begin{algorithm}[b]
		\SetKwInOut{input}{input}
		\SetKwInOut{output}{output}
		\caption{(Learned) D-AMP algorithm \cite{metzler2016denoising,metzler2017learned}}
				\label{algorithmDAMP}
		\input{${\boldsymbol x}_0 = {\boldsymbol 0}, {\boldsymbol y}, {\boldsymbol A}$}
		\For{$t = 1$ to $T$}
		{
			${\boldsymbol b}_t \gets {\boldsymbol z}_{t-1} \mathrm{div} {\boldsymbol D}_{{\boldsymbol w}(\hat{\sigma}_{t-1})}({\boldsymbol x}_{t-1} + {\boldsymbol A}^H {\boldsymbol z}_{t-1}) / M$
			
			${\boldsymbol z}_t \gets {\boldsymbol y} - {\boldsymbol A} {\boldsymbol x}_t + {\boldsymbol b}_t$
			
			$\hat{\sigma}_t \gets \| {\boldsymbol z}_t \|_2 / \sqrt{M}$
			
			${\boldsymbol x}_{t+1} \gets {\boldsymbol D}_{{\boldsymbol w}(\hat{\sigma_t})}({\boldsymbol x}_t + {\boldsymbol A}^H {\boldsymbol z}_t)$ 
		}
		\output{${\boldsymbol x}_T$}
	\end{algorithm}
	
	D-AMP first utilized conventional state-of-the-art denoisers such as BM3D~\cite{dabov2007image} 
	for ${\boldsymbol D}_{{\boldsymbol w}(\hat{\sigma_t})}( \cdot )$ in Algorithm~\ref{algorithmDAMP}~\cite{metzler2015bm3d}.
	Given a standard deviation of noise $\hat{\sigma}_t$ at iteration $t$, BM3D was applied to 
	a noisy image ${\boldsymbol x_t} + {\boldsymbol A^H}{\boldsymbol z_t}$  to yield estimated image ${\boldsymbol x_{t+1}}$.
	Since BM3D can not be represented as a linear function, analytical form for divergence of this denoiser is not available
	for the Onsager term. This issue was resolved by using 
	Monte-Carlo (MC) approximation for divergence term 
	$\mathrm{div} {\boldsymbol D}_{{\boldsymbol w}(\hat{\sigma}_{t})}(\cdot)$~\cite{ramani2008monte}: 
	For $\epsilon>0$,
	\begin{equation} 
	\mathrm{div} {\boldsymbol D}_{{\boldsymbol w}(\hat{\sigma}_{t})}(\cdot) \approx \frac{\tilde{\boldsymbol n}'}{\epsilon}  
	\left( {\boldsymbol D}_{{\boldsymbol w}(\hat{\sigma}_{t})}(\cdot + \epsilon \tilde{\boldsymbol n}) 
	- {\boldsymbol D}_{{\boldsymbol w}(\hat{\sigma}_{t})}(\cdot) \right)
	\label{eq:divD}
	\end{equation} 
	where $\tilde{\boldsymbol n}$ is a standard normal random vector.
	Recently, LDAMP was proposed to use deep learning based denoisers for ${\boldsymbol D}_{{\boldsymbol w}(\hat{\sigma_t})}( \cdot )$ in Algorithm~\ref{algorithmDAMP}~\cite{metzler2017learned}.
	Nine deep neural network denoisers were trained with noiseless ground truth data for different noise levels.
	LDAMP consists of 10 D-AMP layers (iterations) and in each layer, 
	state-of-the-art DnCNN~\cite{zhang2017beyond} was used as a denoiser operator.
	Unlike other unrolled neural network versions of iterative algorithms such as
	Learned-AMP~\cite{borgerding2017amp} and LISTA~\cite{gregor2010learning},
	LDAMP exploited imaging system models and fixed ${\boldsymbol A}$ and ${\boldsymbol A^H}$ operators while the parameters for DnCNN denoisers were trained with ground truth data in image domain.
	
	\subsection{Stein's unbiased risk estimator (SURE) based deep neural network denoisers}
	
	Over the past years, deep neural network based denoisers have been well investigated~\cite{burger2012image, jain2009natural,Vincent:2010vu,xie2012image,zhang2017beyond} and they often outperformed conventional state-of-the-art denoisers such as BM3D.	Deep neural network denoisers such as DnCNN~\cite{zhang2017beyond} yielded state-of-the-art denoising performance at multiple noise levels and are typically trained by minimizing the mean square error (MSE) between the output image of denoiser and the noiseless ground truth image:
	\begin{equation}
	\frac{1}{K}\sum_{j=1}^{K} \| {\boldsymbol D}_{{\boldsymbol w}(\sigma)}( {\boldsymbol z}^{(j)} ) - {\boldsymbol x}^{(j)} \|^2 \label{eq3}
	\end{equation} 
	where ${\boldsymbol z} \in\mathbb{R}^N$ is a noisy image of the ground truth image 
	${\boldsymbol x}$ contaminated with $i.i.d.$ Gaussian noise with zero mean and fixed $\sigma^2$ variance,
	${\boldsymbol D}_{{\boldsymbol w}(\sigma)}( \cdot )$ is a deep learning based denoiser with large-scale parameters ${\boldsymbol w}$ to train,
	and ${({\boldsymbol z}^{(1)}, {\boldsymbol x}^{(1)})}, \ldots, {({\boldsymbol z}^{(K)}, {\boldsymbol x}^{(K)})}$ is a training dataset with $K$ samples
	in image domain.
	
    Recently, a method to train deep learning based denoisers only with noisy images was proposed~\cite{soltanayev2018}.
	Instead of minimizing MSE, the following Monte-Carlo Stein's unbiased risk estimator (MC-SURE) that approximates MSE was minimized with respect to large-scale
	weights in a deep neural network without noiseless ground truth images:
	\begin{equation} 
	\begin{aligned}
	\frac{1}{K}\sum_{j=1}^{K} \|{\boldsymbol z}^{(j)} - {\boldsymbol D}_{{\boldsymbol w}(\sigma)}( {\boldsymbol z}^{(j)} )  \|^2 - N \sigma^2 + \\
	\frac{2\sigma^2 \tilde{\boldsymbol n}'}{\epsilon}  
	\left( {\boldsymbol D}_{{\boldsymbol w}(\sigma)}({\boldsymbol z}^{(j)} + \epsilon \tilde{\boldsymbol n}) 
	- {\boldsymbol D}_{{\boldsymbol w}(\sigma)}({\boldsymbol z}^{(j)}) \right).
	\label{eq4}
	\end{aligned}
	\end{equation} 
	
	In CS image recovery applications, there are often cases where no ground truth data or no Gaussian contaminated image are available, but only CS samples in measurement domain are available for training. Thus, it is not straightforward to use 
	MSE based or MC-SURE based deep denoiser networks for CS image recovery. The goal of this article is to propose a method to train deep neural network denoisers directly from CS samples without image prior and to simultaneously recover high quality images.

	\section{Method}
	
	\subsection{Training deep denoisers from undersampled measurements without ground truth}
	
	Our proposed method exploits D-AMP (LDAMP)~\cite{metzler2016denoising,metzler2017learned} to yield Gaussian noise contaminated images during CS image recovery from large-scale undersampled measurements and to train a single deep neural network denoiser with these noisy images at different noise levels using MC-SURE based denoiser learning~\cite{soltanayev2018}.    Since Onsager correction term in D-AMP allows ${\boldsymbol x}_t + {\boldsymbol A}^H {\boldsymbol z}_t$ term to be close to the ground truth image plus Gaussian noise, so we conjecture that these can be utilized for MC-SURE based denoiser training. We further investigated this in the next subsection.    Our joint algorithm is detailed in Algorithm~\ref{algorithm2}. Note that for large-scale CS measurements 
	${\boldsymbol y}^{(1)}, \ldots, {\boldsymbol y}^{(K)}$, both images $\hat{\boldsymbol x}_{L}^{(1)}, \ldots, \hat{\boldsymbol x}_{L}^{(K)}$ and trained denoising deep network ${\boldsymbol D}_{{\boldsymbol w}_{L}(\sigma)}(\cdot)$ were able to be obtained. After training, fast and high performance CS image recovery was possible without further training of deep denoising network.

	\begin{algorithm}[!t]
		\SetKwInOut{input}{input}
		\SetKwInOut{output}{output}
		\caption{Simultaneous LDAMP and MC-SURE deep denoiser learning algorithm}
				\label{algorithm2}
		\input{${\boldsymbol y}^{(1)}, \ldots, {\boldsymbol y}^{(K)}, {\boldsymbol A}$}
		\For{$l = 1$ to $L$}
		{
			\For{$k = 1$ to $K$}
			{
				
				\For{$t = 1$ to $T$}
				{
					${\boldsymbol b}_t \gets {\boldsymbol z}_{t-1} \mathrm{div} {\boldsymbol D}_{{\boldsymbol w}_{l}(\hat{\sigma}_{t-1})}({\boldsymbol x}_{t-1} + {\boldsymbol A}^H {\boldsymbol z}_{t-1}) / M$
					
					${\boldsymbol z}_t \gets {\boldsymbol y}^{(k)} - {\boldsymbol A} {\boldsymbol x}_t + {\boldsymbol b}_t$
					
					$\hat{\sigma}_t \gets \| {\boldsymbol z}_t \|_2 / \sqrt{M}$

					\eIf{$\hat{\sigma}_t  <= 55 $}{
						${\boldsymbol x}_{t+1} \gets {\boldsymbol D}_{{\boldsymbol w}_{l}(\hat{\sigma_t})}({\boldsymbol x}_t + {\boldsymbol A}^H {\boldsymbol z}_t)$ \
					}{
						${\boldsymbol x}_{t+1} \gets BM3D_{\hat{\sigma_t}}({\boldsymbol x}_t + {\boldsymbol A}^H {\boldsymbol z}_t)$ \
					}
				}
				
				$\hat{\boldsymbol x}_{l}^{(k)} \gets {\boldsymbol x}_{T+1}$
				
				${\boldsymbol s}_l^{(k)}  \gets {\boldsymbol x}_T + {\boldsymbol A}^H {\boldsymbol z}_T$
				
			}
			
			Train ${\boldsymbol D}_{{\boldsymbol w}_{l}(\sigma)}(\cdot)$ with ${\boldsymbol s}_l^{(1)}, \ldots, {\boldsymbol s}_l^{(K)}$ at different noise levels $\sigma$
			
		}
		\output{$\hat{\boldsymbol x}_{L}^{(1)}, \ldots, \hat{\boldsymbol x}_{L}^{(K)}, {\boldsymbol D}_{{\boldsymbol w}_{L}(\sigma)}(\cdot)$ }
	\end{algorithm}
	
    The original LDAMP utilized  9 DnCNN denoisers trained on $``$clean$"$ images for different noise levels ($\sigma = 0-10, 10-20, 20-40, 40-60, 60-80, 80-100, 100-150, 150-300, 300-500$)~\cite{metzler2017learned}. However, in our work we argue that training a single DnCNN denoiser could be enough to achieve almost the same results. The network was pre-trained with reconstructed images using D-AMP with BM3D plus Gaussian noise in $\sigma \in [0, 55]$ range. The pre-trained DnCNN blind denoiser ${\boldsymbol D}_{{\boldsymbol w}_{l}(\hat{\sigma_t})}$ cleans ${\boldsymbol x}_{t-1} + {\boldsymbol A}^H {\boldsymbol z}_{t-1}$ with noise level between $[0, 55]$ (line 8), while $BM3D_{\hat{\sigma_t}}$ is used for higher level noise reduction (line 10). Depending on a sampling ratio and forward operator $\boldsymbol A$, only initial 2-4 iterations are required to use BM3D to decrease the noise level sufficient enough for DnCNN. Then, after $T$ iterations, the set of training data ${\boldsymbol s}_l^{(1)}, \ldots, {\boldsymbol s}_l^{(K)}$ can be generated using LDAMP with pre-trained deep denoiser. Those noisy training images were utilized for further training pre-trained DnCNN with MC-SURE. 

	It is worth to note that the noise level range for DnCNN is subject to change depending on a particular problem. For example, we found that for $i.i.d.$ Gaussian and coded diffraction pattern (CDP) matrices, training DnCNN in $\sigma \in [0, 55]$ range was optimal, while for CS-MRI case, the range was shortened to $\sigma \in [0, 10]$. For noisier measurements, the noise level of the denoiser may need to be optimized.

	\subsection{Accuracy of standard deviation estimation for MC-SURE based denoiser learning}
	
	In D-AMP and LDAMP~\cite{metzler2016denoising,metzler2017learned}, noise level was estimated in measurement domain using
	\begin{equation}
	\hat{\sigma}_t \gets \| {\boldsymbol z}_t \|_2 / \sqrt{M}.
	\label{stdv}
	\end{equation}
	The accuracy of this estimation was not critical for D-AMP or LDAMP since denoisers in both methods were not sensitive to different noise levels.
	However, accurate noise level estimation is quite important for MC-SURE based deep denoiser network learning. We investigated the accuracy of
	(\ref{stdv}). It turned out that the accuracy of noise level estimation depends on measurement matrices.
		
	With $i.i.d.$ Gaussian measurement matrix ${\boldsymbol A}$, (\ref{stdv}) was very accurate and comparable to the ground truth standard deviation 
	that was obtained from the true residual $({\boldsymbol x_t}+{\boldsymbol A}^H {\boldsymbol z_t}) - {\boldsymbol x_{true}}$. However, with 
	CDP measurement matrix ${\boldsymbol A}$ that yields complex measurements, it turned out that (\ref{stdv}) yielded over-estimated
	noise level for multiple examples.
	Since the image ${\boldsymbol x_t}$ is real, we propose a new standard deviation estimation method for D-AMP:
	\begin{equation}
	\hat{\sigma}_t \gets \| \operatorname{Re}({\boldsymbol A}^H {\boldsymbol z}_t ) \|_2 / \sqrt{N}.
	\label{new_sigma}
	\end{equation}
	We performed comparison studies among (\ref{stdv}), (\ref{new_sigma}), and the ground truth from true residual 
	$({\boldsymbol x_t}+{\boldsymbol A}^H {\boldsymbol z_t}) - {\boldsymbol x_{true}}$ and found that
	they are all similar for $i.i.d.$ Gaussian measurement matrix. However, our proposed method  (\ref{new_sigma}) yielded more accurate estimates
	of standard deviation than previous method (\ref{stdv}) for CDP matrix with complex numbers.

	\begin{figure}[!t]
		\begin{center}
			\begin{subfigure}[t]{0.48\linewidth}
				\centering
				\scriptsize (a) True residual
				\includegraphics[width=1\textwidth]{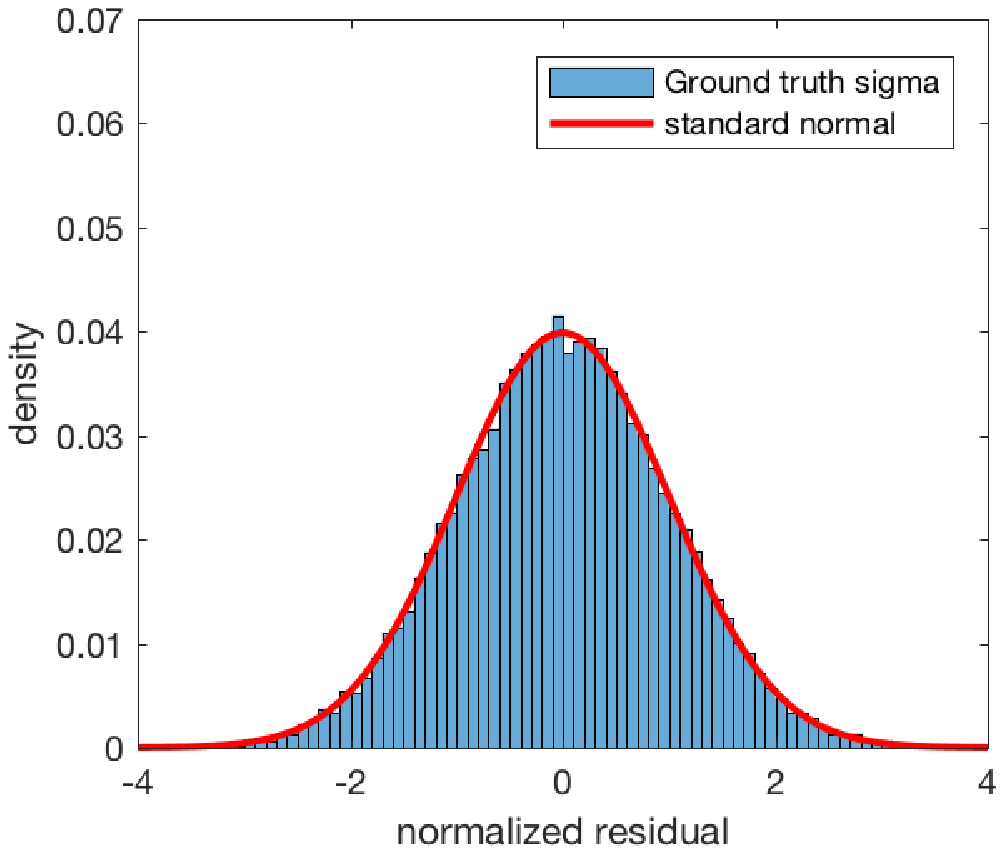}  \label{sfig1}
			\end{subfigure}
			\begin{subfigure}[t]{0.48\linewidth}
				\centering
				\scriptsize (b) D-AMP
				\includegraphics[width=1\textwidth]{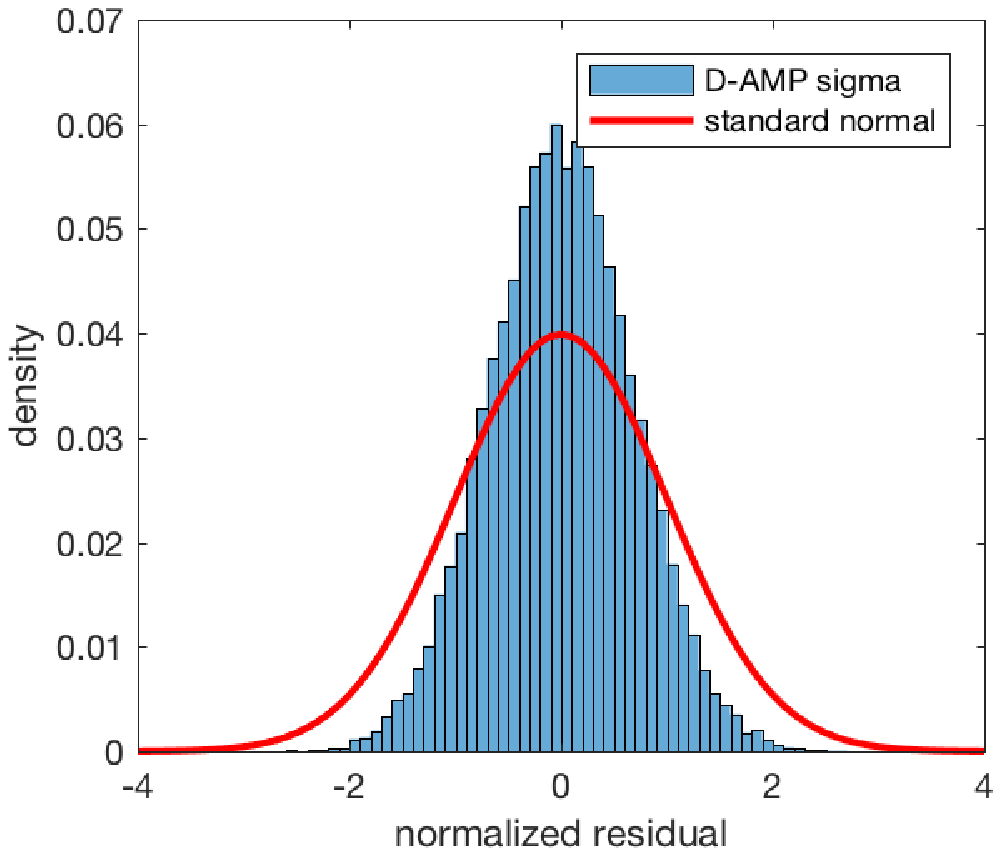} \label{sfig2}
			\end{subfigure}
			\begin{subfigure}[t]{0.48\linewidth}
				\centering
				\scriptsize (c) Proposed
				\includegraphics[width=1\textwidth]{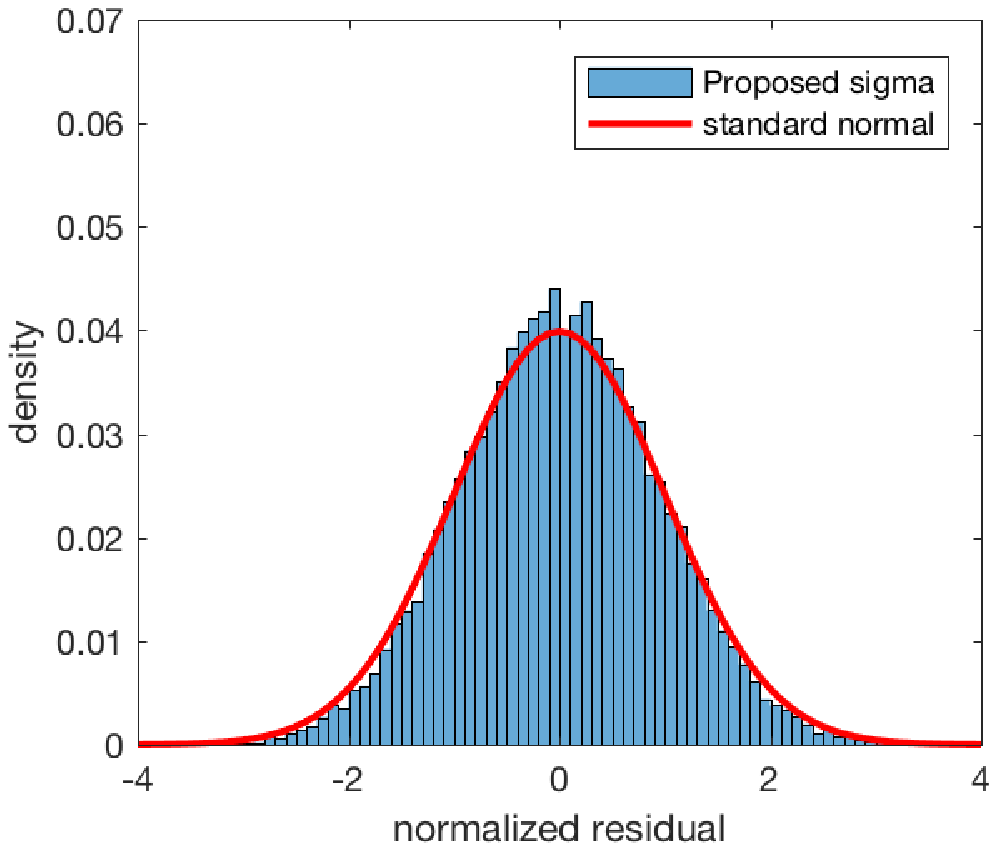}  \label{sfig3}
			\end{subfigure}
			\begin{subfigure}[t]{0.48\linewidth}
				\begin{minipage}{.1cm}
            				\vfill
            			\end{minipage}			
            		\end{subfigure}
		\caption{Normalized residual histograms of ``Boat" image after 10 iterations using LDAMP-BM3D 
		for CDP matrix. Normalization was done with estimated sigma from (a) true residual (b) ${\boldsymbol z}_T$ (D-AMP) and (c) $\operatorname{Re}({\boldsymbol A}^H {\boldsymbol z}_T )$ (Proposed).}
		\label{PDF}
		\end{center}
	\end{figure}

	Figure~\ref{PDF} illustrates the accuracy of our noise estimator compared to the previous one. 
	Normalizing true residual histogram with the true noise level yielded good fitting to standard normal density (red line) as shown in Figure~\ref{PDF}(a).
	Normalized histogram of true residual with the previous noise estimation method yielded sharper histogram as shown in Figure~\ref{PDF}(b) due to the overestimation of noise levels. However, our proposed standard deviation estimation yielded good normalized histogram fitting to the ground truth (red line).	
	In the simulation, we will show that our proposed estimation is critical to achieve high performance using our proposed methods with CDP and CS-MRI.
	
	Moreover, we found out that proposed noise estimator (\ref{new_sigma}) can also be applicable for CS-MRI cases, when k-space data is moderately undersampled. 
	Therefore, for a sampling rate of larger than 35-40\%, true residual follows a Gaussian noise, which can be accurately measured by (\ref{new_sigma}) and further utilized for training deep denoisers with MC-SURE.

	\section{Simulation Results}
	
	\subsection{Setup}
	
	\paragraph{Datasets}
	 We used images from DIV2K~\cite{agustsson2017ntire}, Berkeley's BSD-500~\cite{MartinFTM01} datasets, and standard test images for training and testing our proposed method on $i.i.d.$ Gaussian and CDP matrices. Training dataset was comprised of all 500 images from BSD-500, while a test set of 100 images included 75 randomly chosen images from DIV2K and 25 standard test images. Since the proposed method uses fixed linear operator for image reconstruction, all test and train images had the same size. Thus, all images were subsampled and cropped to a size of 180 $\times$180 and compressively sampled to generate measurement data.
	
	For CS-MRI reconstruction, training data were pulled from an open repository (http://mridata.org/). 
	The knee dataset includes 256 slices of 320 $\times$ 320 knee images per patient for 20 patients.	
	We chose images from 3 patients randomly for training and other images from 1 patient for testing.	
	The images were transformed onto the k-space domain and subsampled with realistic radial sampling patterns at various sampling rates. 
	
	We implemented all methods on the Tensorflow framework and used Adam optimizer~\cite{kinga2015method} with the learning rate of 10\textsuperscript{-3}, which was dropped to 10\textsuperscript{-4} after 40 epochs and further trained for 10 epochs. The batch size was set to 128 and training DnCNN denoiser took approximately 12-14 hours on an NVIDIA Titan Xp.     
	
	\paragraph{Initialization of DnCNN denoiser}
	The primary idea of this stage is to decouple DnCNN denoiser from the LDAMP SURE and pre-train it given measurement data $\boldsymbol y$ and linear operator $\boldsymbol A$. To do so, BSD-500 images were firstly reconstructed using BM3D-AMP. Recovered images were rescaled, cropped, flipped, and rotated to generate 298,000 50$\times$50 patches. These patches were used as ground truth to pre-train DnCNN denoiser with MSE. We also simulated another scenario. Since our approach does not require dataset with ground truth, it is possible to use measurement data from the test set. Thus, we generated 357,600 50$\times$50 patches from reconstructed test and train images. Pre-trained DnCNN denoiser on those patches was utilized in LDAMP network and the network was labeled in tables as  "LDAMP BM3D-T" or "LDAMP BM3D" depending on whether test measurements were included for training or not. Our DnCNN denoiser was trained for $\sigma \in [0, 55]$ noise level range. 
	
	In the CS-MRI case, BM3D-AMP-MRI  \cite{eksioglu2018denoising} was specifically tailored for CS-MRI reconstruction and thus yielded significantly better results than conventional BM3D-AMP. Therefore, k-space knee dataset was reconstructed using BM3D-AMP-MRI \cite{eksioglu2018denoising} and resulted images were rescaled, cropped, flipped, and rotated to generate 267,240 50$\times$50 patches for LDAMP BM3D and 350,320 50$\times$50 patches for LDAMP BM3D-T training. We trained DnCNN denoisers for $\sigma \in [0, 10]$ noise range. 
	
	\paragraph{Training LDAMP SURE}
	Firstly, LDAMP SURE was run $T=10$ iterations using pre-trained DnCNN denoiser and BM3D. At the last iteration, we collected images and estimated noise standard deviation with (\ref{new_sigma}). Then, all images with noise levels in $[0, 55]$ range (CS-MRI case: $\sigma \in [0, 10]$) were grouped into one set, while outliers that have larger noise levels were replaced by Gaussian noise added BM3D-AMP recovered images. Overall, we have dataset of all images with $\sigma \in [0, 55]$ (CS-MRI case: $\sigma \in [0, 10]$) to train DnCNN denoiser with MC-SURE. These steps were repeated $L$ times to further improve the performance of our proposed method. 
	
	Although training DnCNN with MC-SURE involves estimation of a noise standard deviation for an entire image, we assume that a patch from an image has the same noise level as the image itself. Thus, we generated patches without using rescaling to avoid noise distortion to train LDAMP SURE.
	
	To train DnCNN with SURE, we used weights of pre-trained DnCNN and trained using Adam optimizer \cite{kinga2015method} with the learning rate of 10\textsuperscript{-4} and batch size 128 for 10 epochs. Then, we decreased learning rate to 10\textsuperscript{-5} and trained for 10 epochs. Training process lasted about 3-4 hours for LDAMP SURE or LDAMP SURE-T respectively. We empirically found that after $L$=2 iterations (Algorithm~\ref{algorithm2}-line 1) of training LDAMP SURE, the results converge for both CDP and $i.i.d.$ Gaussian cases, while for CS-MRI: $L$ = 1.      
	
	The accuracy of MC-SURE approximation depends on the selection of constant value $\epsilon$, which is directly proportional to $\sigma$ \cite{deledalle2014stein, soltanayev2018}. Therefore, for training DnCNN with SURE, $\epsilon$ value was calculated for each patch based on its noise level. 
	
	\subsection{Results}
	
	\paragraph{Gaussian measurement matrix}
	    We compared our proposed LDAMP SURE with the state-of-the-art CS methods, namely BM3D-AMP\cite{metzler2015bm3d}, NLR-CS\cite{dong2014compressive}, and TVAL3\cite{li2009user}. BM3D-AMP was used with default parameters and run for 30 iterations to reduce high variation in the results, although 
	PSNR\footnote{PSNR stands for peak signal-to-noise ratio and is calculated by following expression: $10log_{10}(\frac{255^2}{mean(\hat{x} - x_{gt})^2})$ for pixel range $\in [0-255]$} 
	approached its maximum after 10 iterations \cite{metzler2016denoising}. The proposed LDAMP SURE algorithm was run 30 iterations but also showed convergence after 8-10 iterations. NLR-CS was initialized with 8 iterations of BM3D as justified in \cite{metzler2016denoising}, while TVAL3 was set to its default parameters. Also, we included the results of LDAMP trained on ground truth images to see the performance gap. 
	
\begin{figure*}[!t]
		\begin{center}
			\begin{subfigure}[!b]{0.16\textwidth}
				\centering
				Ground truth
				\includegraphics[width=1\textwidth]{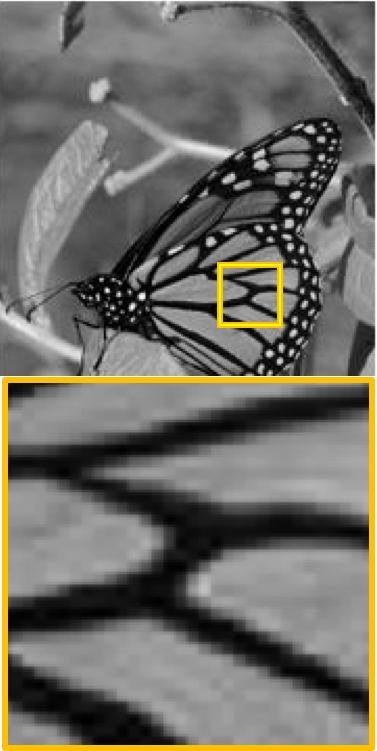}
				\caption{PSNR}
			\end{subfigure}
			\begin{subfigure}[!b]{0.16\textwidth}
				\centering
				TVAL3 
				\includegraphics[width=1\textwidth]{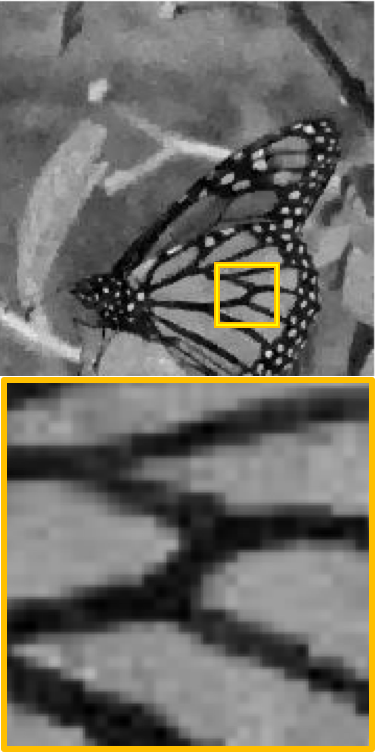}
				\caption{26.10 dB}
			\end{subfigure}
			\begin{subfigure}[!b]{0.16\textwidth}
				\centering
				BM3D-AMP 
				\includegraphics[width=1\textwidth]{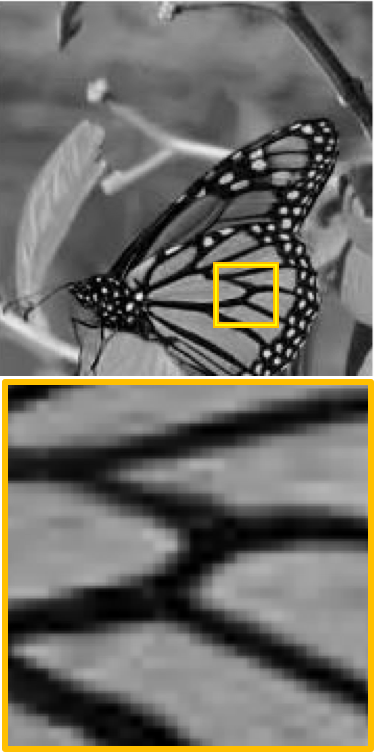}
				\caption{31.90 dB}
			\end{subfigure}
			\begin{subfigure}[!b]{0.16\textwidth}
				\centering
				NLR-CS 
				\includegraphics[width=1\textwidth]{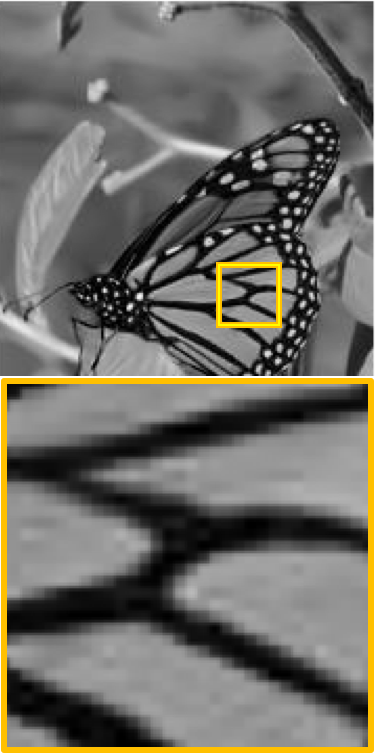}
				\caption{33.59 dB}
			\end{subfigure}
			\begin{subfigure}[!b]{0.16\textwidth}
				\centering
				LDAMP SURE 
				\includegraphics[width=1\textwidth]{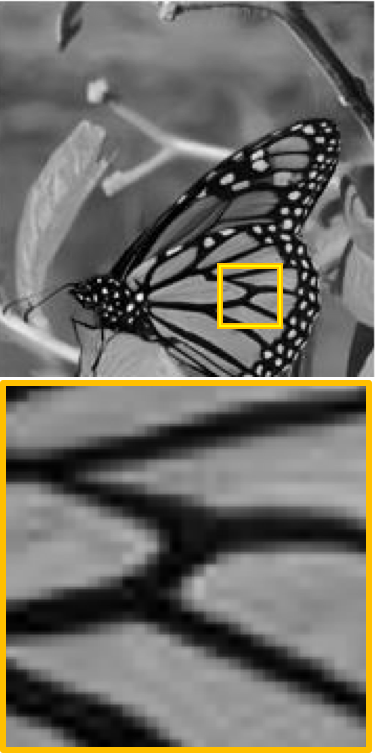}
				\caption{34.53dB}
			\end{subfigure}
			\begin{subfigure}[!b]{0.16\textwidth}
				\centering
				LDAMP SURE-T 
				\includegraphics[width=1\textwidth]{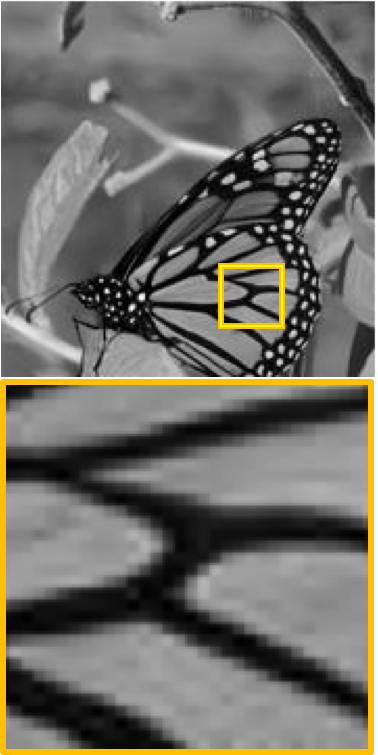}
				\caption{\textbf{35.19 dB}}
			\end{subfigure}
		\caption{Reconstructions of 180$\times$180 test $``$Butterfly$"$ image with i.i.d. Gaussian matrix with ${M}/{N}=0.25$ sampling rate.}
		\label{fig-Gaussian_recon}
		\end{center}
	\end{figure*}

	From Table \ref{gaussian}, proposed LDAMP SURE and LDAMP SURE-T outmatches other methods at higher CS ratios by 0.26-0.46 dB, while at a highly undersampled case, it is inferior to NLR-CS. Nevertheless, it is clear that SURE based LDAMP is able to improve the performance of pretrained LDAMP BM3D and surpasses BM3D-AMP by 0.28-1.56 dB. In Figure~\ref{fig-Gaussian_recon}, reconstructions of all methods on a test image are represented for 25\% sampling ratio. Proposed LDAMP SURE and LDAMP SURE-T provide sharper edges and preserve more details.     
	
	In terms of run time, the dominant source of computation comes from using BM3D denoiser at initial iterations, while DnCNN takes less than a second for inference. LDAMP SURE utilizes CPU for BM3D and GPU for DnCNN. Consequently, proposed LDAMP SURE is comparatively faster than BM3D-AMP, NLR-CS, and TVAL3 methods.  
	
\begin{figure*}[!t]
		\begin{center}
			\begin{subfigure}[!b]{0.16\textwidth}
				\centering
				Ground truth
				\includegraphics[width=1\textwidth]{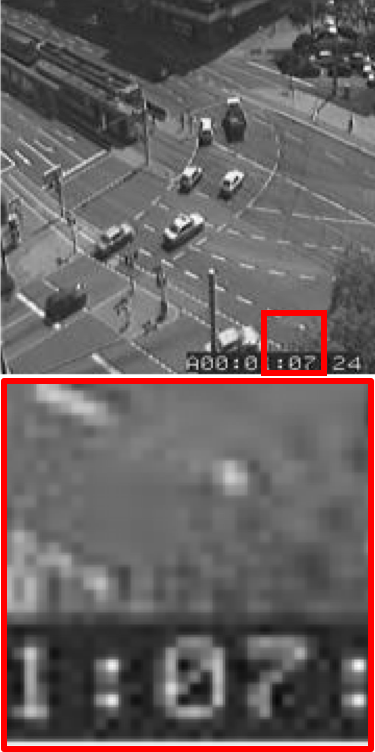}
				\caption{PSNR}
			\end{subfigure}
			\begin{subfigure}[!b]{0.16\textwidth}
				\centering
				TVAL3 
				\includegraphics[width=1\textwidth]{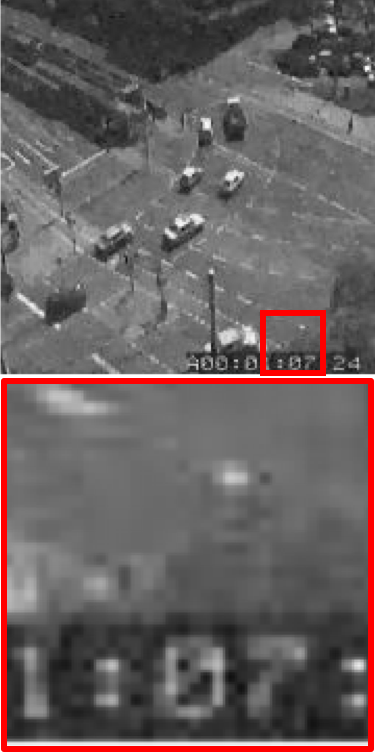}
				\caption{27.52 dB}
			\end{subfigure}
			\begin{subfigure}[!b]{0.16\textwidth}
				\centering
				BM3D-AMP 
				\includegraphics[width=1\textwidth]{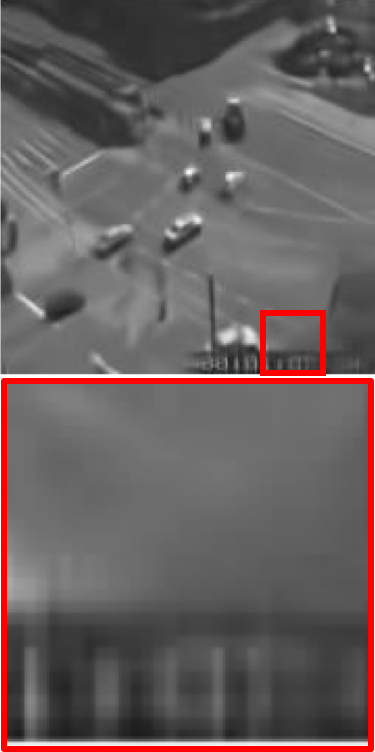}
				\caption{24.08 dB}
			\end{subfigure}
			\begin{subfigure}[!b]{0.16\textwidth}
				\centering
				NLR-CS 
				\includegraphics[width=1\textwidth]{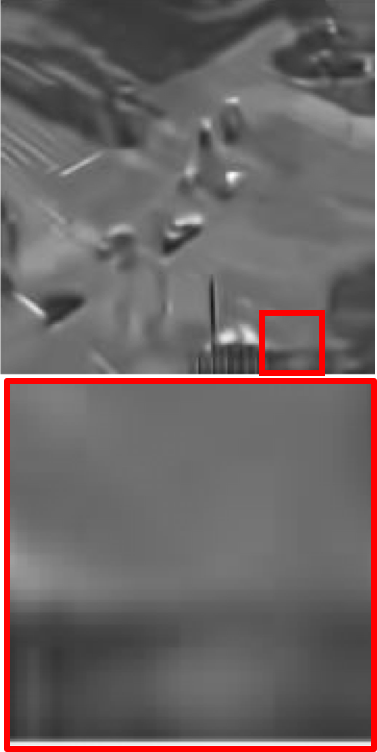}
				\caption{22.29 dB}
			\end{subfigure}
			\begin{subfigure}[!b]{0.16\textwidth}
				\centering
				LDAMP SURE 
				\includegraphics[width=1\textwidth]{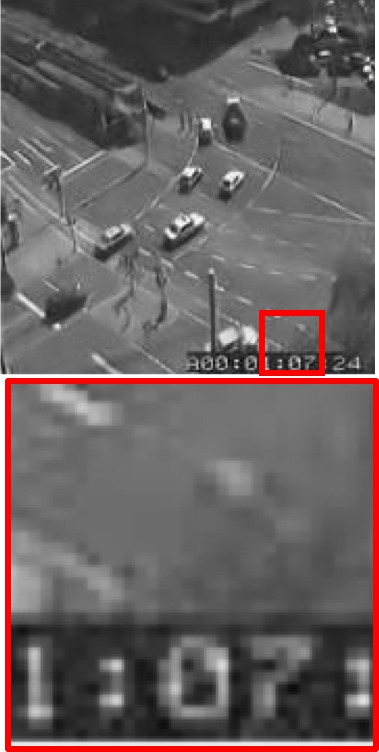}
				\caption{\textbf{29.17 dB}}
			\end{subfigure}
			\begin{subfigure}[!b]{0.16\textwidth}
				\centering
				LDAMP SURE-T 
				\includegraphics[width=1\textwidth]{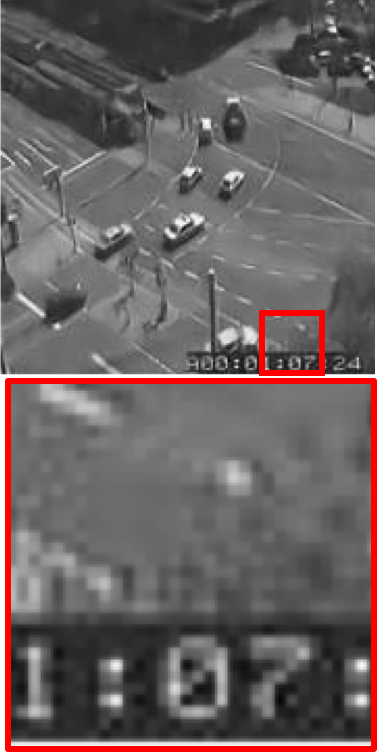}
				\caption{28.92 dB}
			\end{subfigure}
		\caption{Reconstructions of 180$\times$180 test image with CDP measurement matrix for ${M}/{N}=0.15$ sampling rate.}
		\label{fig-CDP_recon}
		\end{center}
	\end{figure*}

	\paragraph{Coded diffraction pattern measurements} 
    LDAMP SURE was tested with randomly sampled coded diffraction pattern \cite{candes2015phase} and yielded the best quantitative performance at higher sampling rates (see Table  \ref{cdp} and Figure \ref{fig-CDP_recon}). LDAMP SURE and LDAMP SURE-T achieved about 1.8 dB performance gain over BM3D-AMP. However, at extremely low sampling ratio, our method slightly falls behind TVAL3. LDAMP SURE requires better dataset than BM3D-AMP reconstructed images from highly undersampled data to pretrain DnCNN. Therefore, one way to surpass TVAL3 at the highly undersampled case is to pretrain DnCNN with TVAL3 reconstructed images. 

\begin{figure*}[!t]
	\begin{center}
		\begin{subfigure}[!b]{0.16\textwidth}
			\centering
			Ground truth
			\includegraphics[width=1\textwidth]{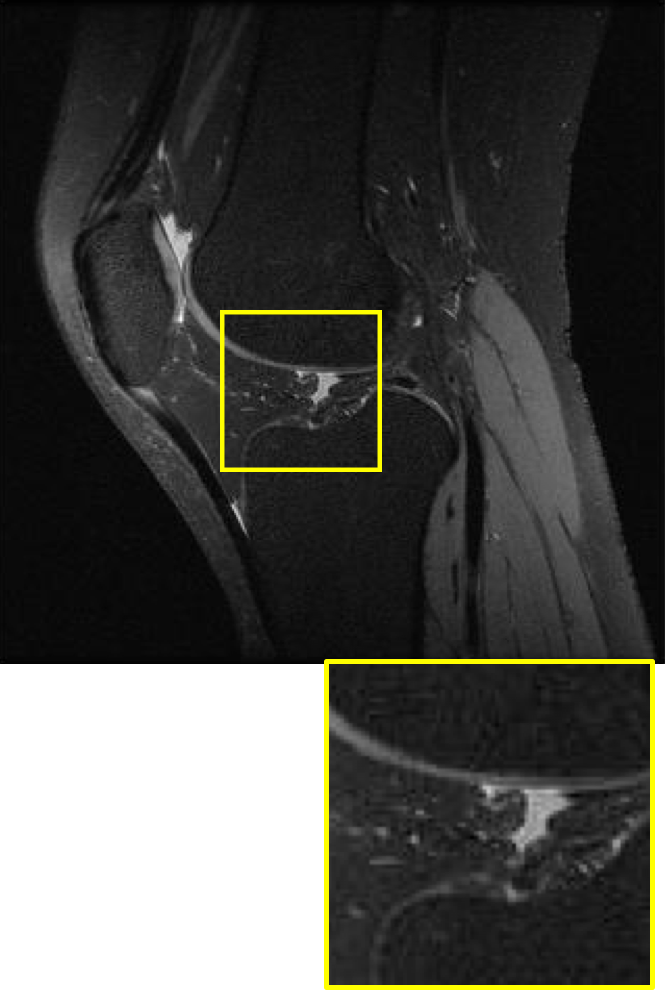}
			\caption{PSNR}
		\end{subfigure}
		\begin{subfigure}[!b]{0.16\textwidth}
			\centering
			TVAL3 
			\includegraphics[width=1\textwidth]{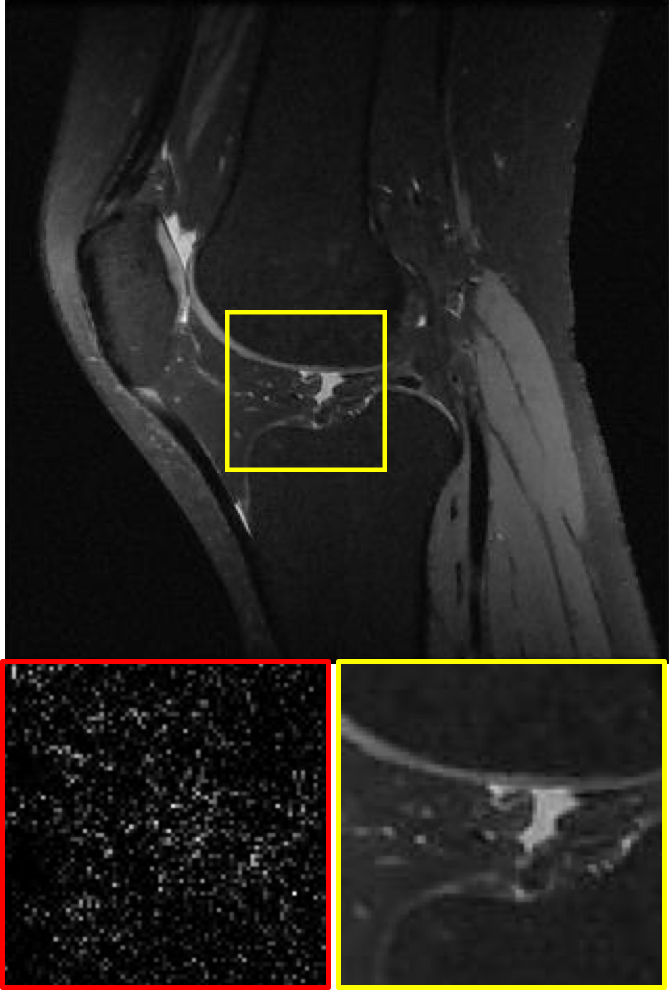}
			\caption{37.44 dB}
		\end{subfigure}
		\begin{subfigure}[!b]{0.16\textwidth}
			\centering
			BM3D-AMP 
			\includegraphics[width=1\textwidth]{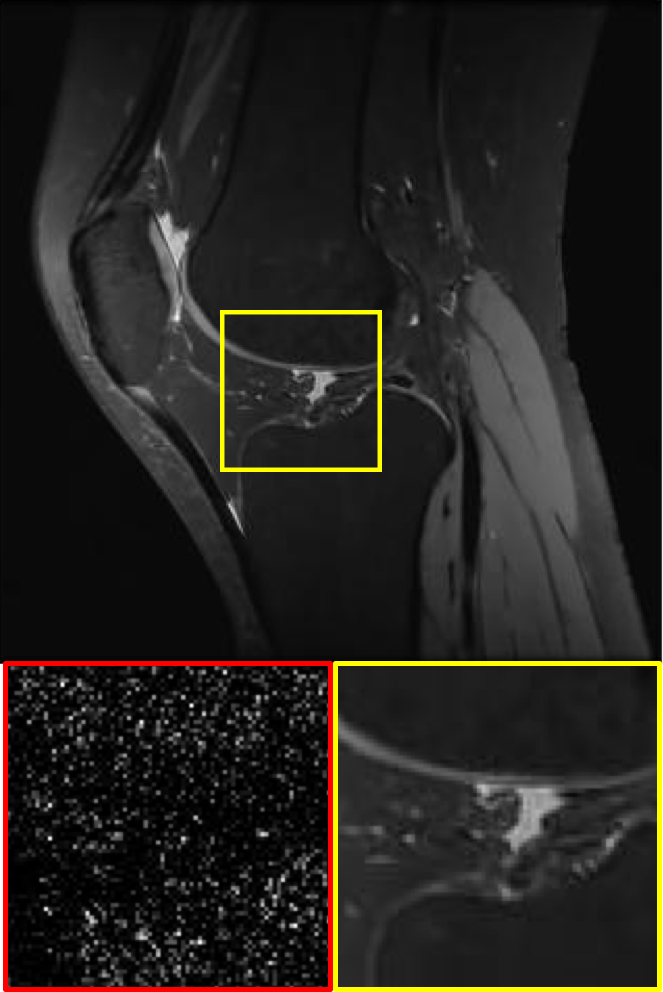}
			\caption{36.54 dB}
		\end{subfigure}
		\begin{subfigure}[!b]{0.16\textwidth}
			\centering
			DL-MRI
			\includegraphics[width=1\textwidth]{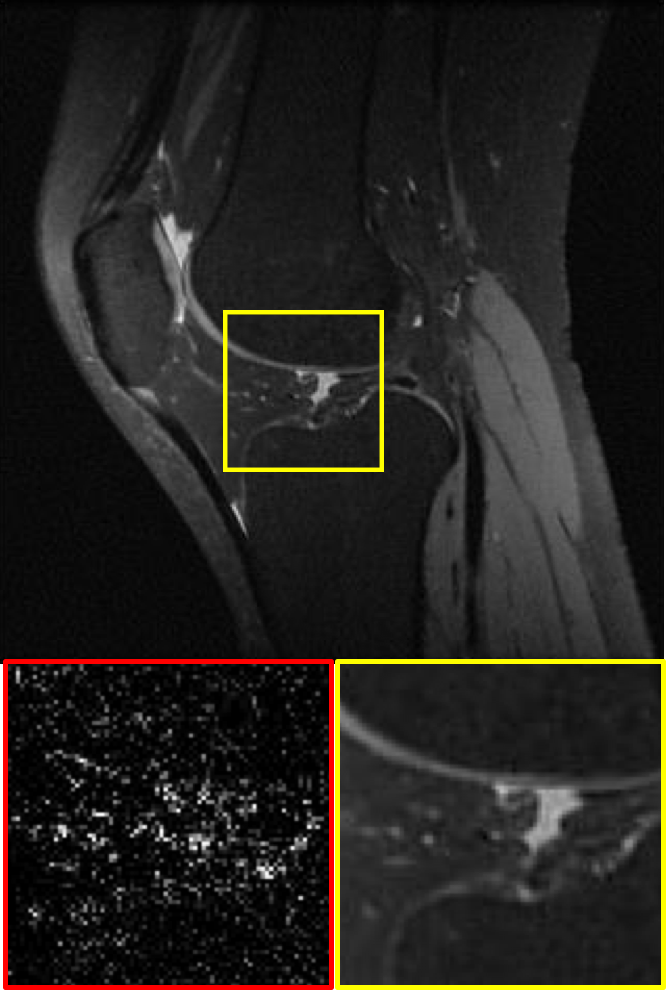}
			\caption{36.76 dB}
		\end{subfigure}
		\begin{subfigure}[!b]{0.16\textwidth}
			\centering
			BM3D-AMP-MRI 
			\includegraphics[width=1\textwidth]{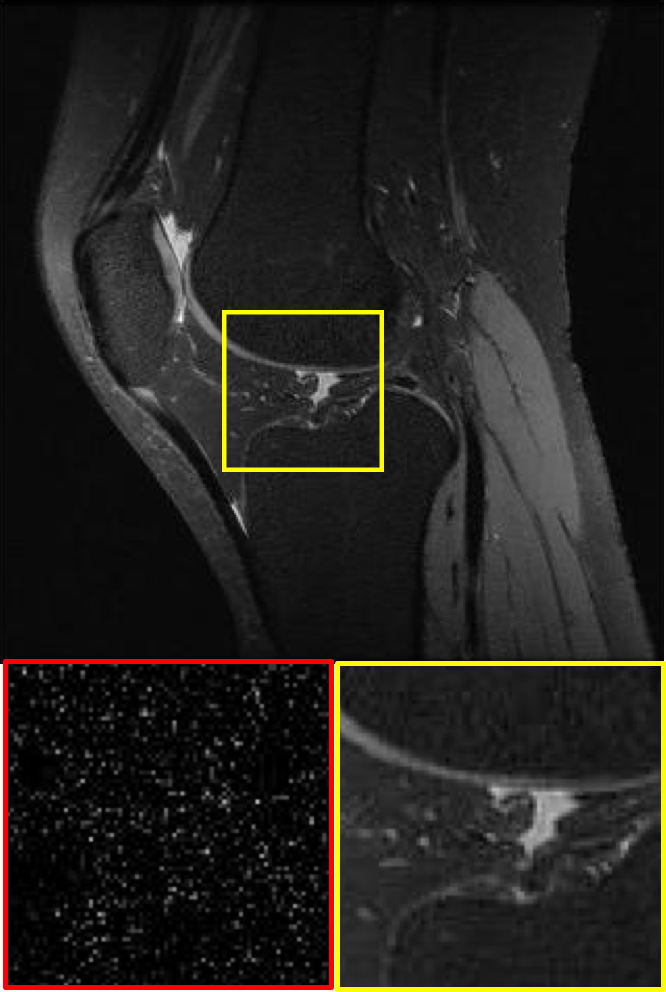}
			\caption{37.85 dB}
		\end{subfigure}
		\begin{subfigure}[!b]{0.16\textwidth}
			\centering
			LDAMP SURE-T 
			\includegraphics[width=1\textwidth]{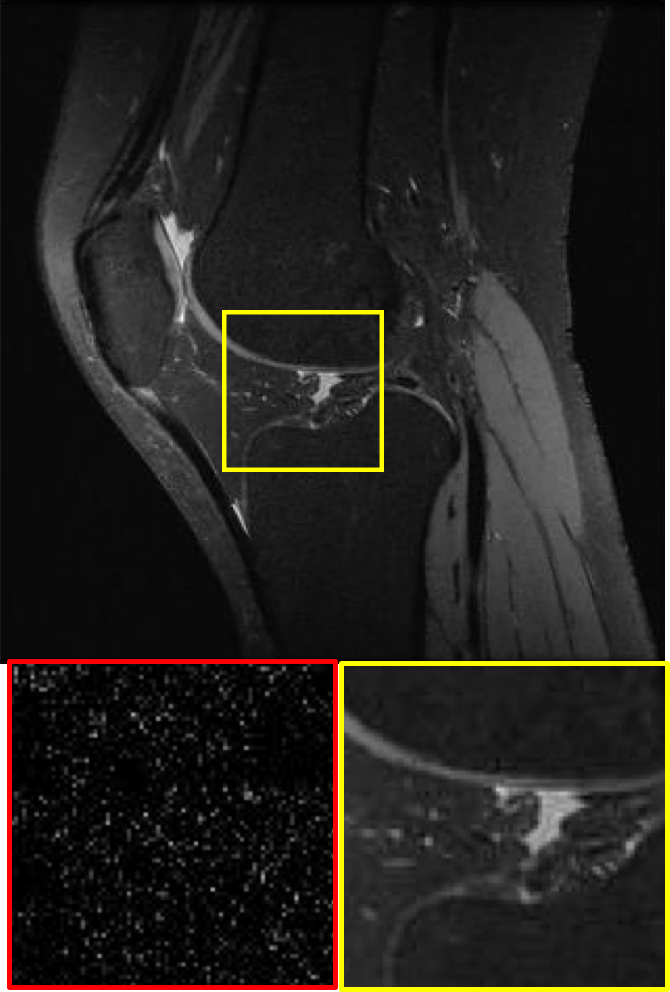}
			\caption{\textbf{38.22 dB}}
		\end{subfigure}
	\caption{Reconstructions with CS-MRI measurement matrix for ${M}/{N}=0.40$. Residual errors are shown in red boxes. }
	\label{fig-CSMRI_recon40}
	\end{center}
	\end{figure*}

	\paragraph{CS MR measurement matrix}  
	LDAMP SURE was applied to CS-MRI reconstruction problem to demonstrate its generality and to show its performance on images that contain structures different from natural image dataset.
	We compared LDAMP SURE with state-of-the-art BM3D-AMP-MRI algorithm \cite{eksioglu2018denoising}  for CS-MR image reconstruction along with TVAL3, BM3D-AMP, and dictionary learning method or DL-MRI \cite{ravishankar2011mr}. Average image recovery PSNRs and run times are tabulated in Table \ref{csmri}.  Figure~\ref{fig-CSMRI_recon40} shows that our proposed method yielded state-of-the-art performance, close to the ground truth.
	The results reveal that proposed LDAMP SURE-T outperforms existing algorithms in all sampling ratios. 
	
	\section{Discussion and Conclusion}
	We proposed simultaneous compressive image recovery and deep learning based denoiser learning method from undersampled measurements. Our proposed method yielded better image quality than conventional methods at higher sampling rates for $i.i.d$ Gaussian, CDP, and CS MR measurements. Thus, it may be possible that this work can be helpful for areas where obtaining ground truth images is challenging such as hyperspectral or medical imaging. 
	
	Note that training deep learning based image denoisers from undersampled data still requires to contain enough information in the undersampled measurements. Tables~\ref{gaussian} and \ref{cdp} show that 5\% of the full samples is not enough to achieve state-of-the-art performance possibly due to lack of information in the measurement.
	Also note that since we assume a single CS measurement for each image and evaluated with various CS matrices with high-resolution images, it was not possible to compare our method to noise2noise~\cite{Lehtinen:2018un} and AmbientGAN~\cite{Bora:2017tz,Bora:2018tl}. Lastly, our proposed method can potentially be used with more advanced deep denoisers as far as they are trainable with MC SURE loss~\cite{soltanayev2018}. 

	\begin{table*}[t]
		\begin{center}
			\begin{tabular} {lccccccc} 
				\toprule
				\multirow{2}{*}{Method} & \multirow{2}{*}{Training Time} & \multicolumn{2}{c}{${M}/{N} = 5\%$}  & \multicolumn{2}{c}{${M}/{N}  = 15\%$}  & \multicolumn{2}{c}{${M}/{N}  = 25\%$}         \\
				\cmidrule(r){3-8}
				& & PSNR & Time & PSNR & Time & PSNR & Time \\
				\midrule
				TVAL3 	   &N/A		&20.46          &9.71     &24.14  	&22.96    	&26.77    	&34.87  \\
				NLR-CS 	   &N/A		&\textbf{21.88}	&128.73	  &27.58	&312.92  	&31.20		&452.23 \\
				BM3D-AMP   &N/A		&21.40    		&25.98 	  & 26.74   &24.21    	& 30.10    	&23.08  \\
				\midrule
				LDAMP BM3D       &10.90 hrs	  &21.41    &8.98    & 27.54   &3.94    & 31.20     &2.89 \\
				LDAMP BM3D-T   &14.30 hrs	&21.42   &8.98    & 27.61   &3.94    & 31.32     &2.89 \\
				LDAMP SURE       &15.05 hrs	  &21.44   &8.98    & 27.65   &3.94    & 31.46     &2.89 \\
				LDAMP SURE-T    &17.97 hrs	 &21.68   &\textbf{8.98} &\textbf{27.84} &\textbf{3.94} &\textbf{31.66}    &\textbf{2.89}\\
				\midrule
				LDAMP MSE 		& 10.17 hrs 	&22.07  &8.98 	&27.78	&3.94	&31.65	&2.89 \\
				\bottomrule
			\end{tabular}
		\caption{Average PSNRs (dB) and run times (sec) of 100 180$\times$180 image reconstructions for i.i.d. Gaussian measurements case (no measurement noise) at various sampling rates ($M/N\times100\%$).}
		\label{gaussian}
		\end{center}
	\end{table*}
	\begin{table*}[!h]
		\begin{center}
			\begin{tabular} {lccccccc} 
				\toprule
				\multirow{2}{*}{Method} & \multirow{2}{*}{Training Time} & \multicolumn{2}{c}{${M}/{N} = 5\%$}  & \multicolumn{2}{c}{${M}/{N} = 15\%$}  & \multicolumn{2}{c}{${M}/{N} = 25\%$}         \\
				\cmidrule(r){3-8}
				& & PSNR & Time & PSNR & Time & PSNR & Time \\
				\midrule
				TVAL3 				 &N/A 	 &\textbf{22.57} 	&\textbf{0.85}	&27.99		&\textbf{0.75}	  &32.82	  &\textbf{0.67}  \\
				NLR-CS 				&N/A 	&19.00  				&93.05 	    	  &22.98	  &86.90  		  	  &31.24	   &119.70\\
				BM3D-AMP         &N/A 	 &21.66    				&22.15   	 	   & 27.29 	   &22.28  		       &31.40   	&17.00 \\
				\midrule
				LDAMP BM3D         &10.56 hrs   &21.97   &23.43    &28.04          &7.01    &31.65    	       &2.71   \\
				LDAMP BM3D-T  	 &12.67 hrs   &21.93   &23.43    &28.01          &7.01    &32.12    	       &2.71   \\
				LDAMP SURE         &15.22 hrs   &22.18   &23.43    &29.14          &7.01    &33.26    	       &2.71   \\
				LDAMP SURE-T      &17.61 hrs   &22.06   &23.43    &\textbf{29.17} &7.01    &\textbf{33.51}    &2.71   \\
				\midrule
				LDAMP MSE		  & 10.17 hrs &22.12   &23.43    &28.87		 &7.01		&33.88	 &2.71 \\
				\bottomrule
			\end{tabular}
		\caption{Average PSNRs (dB) and run times (sec) of 100 180x180 image reconstructions for CDP measurements case (no measurement noise) at various sampling rates ($M/N \times100\%$).}
		\label{cdp}
		\end{center}
	\end{table*}
	\begin{table*}[!h]
		\begin{center}
			\begin{tabular} {lccccccc} 
				\toprule
				\multirow{2}{*}{Method} & \multirow{2}{*}{Training Time} & \multicolumn{2}{c}{${M}/{N}  = 40\%$}  & \multicolumn{2}{c}{${M}/{N}  = 50\%$}  & \multicolumn{2}{c}{${M}/{N}  = 60\%$}         \\
				\cmidrule(r){3-8}
				& & PSNR & Time & PSNR & Time & PSNR & Time \\
				\midrule
				TVAL3 			& N/A 	&36.76	&\textbf{0.58}	  &37.13	&\textbf{0.24}	&38.35	&\textbf{0.21} \\
				DL-MRI			& N/A	&36.60 	&98.51			  &37.81  	&97.58 		    &39.13	&99.44\\
				BM3D-AMP-MRI    & N/A 	&37.42 	&14.76 			  &38.94  	&15.00 		    &40.51	&15.36 \\
				BM3D-AMP        & N/A   &36.15  &96.23  		  &36.29  	&84.34 			&39.53  &98.01 \\
				\midrule
				LDAMP BM3D		& 9.31 hrs   &37.12            &6.26   &38.63           &6.14  &39.53               &6.10  \\
				LDAMP BM3D-T  	&12.41 hrs   &37.65            &6.26   &38.92           &6.14  &39.87 	            &6.10  \\
				LDAMP SURE      &12.04 hrs   &37.40            &6.26   &38.70	        &6.14  &40.62               &6.10  \\
				LDAMP SURE-T    & 16.05 hrs  &\textbf{37.77}   &6.26   &\textbf{39.09}  &6.14  & \textbf{40.71}     &6.10  \\
				\bottomrule
			\end{tabular}
		\caption{Average PSNRs (dB) and run times (sec) of 100 180x180 image reconstructions for CS-MRI measurements case (no measurement noise) at various sampling rates ($M/N \times100\%$).}
		\label{csmri}		
		\end{center}
	\end{table*}	
	

\section*{Acknowledgments}

This work was supported by 
Basic Science Research Program through the National Research Foundation of Korea(NRF) 
funded by the Ministry of Education(NRF-2017R1D1A1B05035810).

	{\small
		\bibliographystyle{ieee}
		\bibliography{arxiv_cstrain}
	}
\end{document}